\documentclass[11pt]{article}

\usepackage[final]{acl}

\usepackage{wrapfig}
\usepackage{subcaption}
\usepackage{times}
\usepackage{latexsym}
\usepackage{overpic}
\usepackage{amsmath}
\usepackage{algorithm}
\usepackage{algpseudocode}
\usepackage{booktabs}
\usepackage{multirow}
\usepackage{amsmath}
\usepackage{amssymb}
\usepackage{colortbl}
\usepackage{xcolor}

\definecolor{mygray}{gray}{0.9}
\definecolor{darkpink}{rgb}{0.91, 0.33, 0.5}
\newcommand{\ourmethodabbr}{PAPO} %
\newcommand{\ourmethod}{Process Aligned Policy Optimization}

\usepackage[T1]{fontenc}

\usepackage[utf8]{inputenc}

\usepackage{microtype}

\usepackage{inconsolata}

\usepackage{graphicx}

\title{Back on Track: Aligning Rewards and States for Reasoning \\in Diffusion Large Language Models}

\author{Yawen Shao$^{1,2}$\thanks{~Work done at Tongyi Lab.}, Jie Xiao$^{2}$, Kai Zhu$^{2}$\thanks{~Corresponding author.}, Yu Liu$^{2}$,  Hongchen Luo$^{3}$, \\ \textbf{Xueyang Fu}$^{1}$, \textbf{Yang Cao}$^{1}$, \textbf{Wei Zhai}$^{1}$\footnotemark[2], \textbf{Zheng-Jun Zha}$^{1}$\\
{$^{1}$~University of Science and Technology of China} \\
{$^{2}$~Tongyi Lab, Alibaba Group} \qquad
{$^{3}$~Northeastern University} \\
\small{\texttt{shaoyawen@mail.ustc.edu.cn}} \qquad
\small{\texttt{\{zkzy@mail, wzhai056@\}ustc.edu.cn}}
}

\begin{document}
\maketitle
\begin{abstract}
Reinforcement learning (RL) holds immense promise for enhancing the reasoning capabilities of diffusion large language models (dLLMs). However, progress is fundamentally constrained by a dual misalignment between authentic generation trajectory and the gradient update process: (i) Process-reward misalignment. Sparse, terminal rewards are indiscriminately assigned to all intermediate steps of the generation process, failing to provide discriminative credit assignment. (ii) State-trajectory misalignment. Policy updates are often diverted toward artificial, out-of-trajectory states, squandering gradients on less informative samples. To address these limitations, we introduce Process Aligned Policy Optimization (PAPO), a novel framework that holistically aligns the RL update with the dLLM's generative trajectory via Step-Aware Process Rewards (SPR) that transform sparse terminal rewards into dense, step-wise credit, and Entropy-Guided Historical Re-enactment (EHR) that replays authentic trajectories at high-uncertainty steps. Extensive experiments on four benchmarks demonstrate that PAPO significantly outperforms baselines, achieving gains of \textbf{4.5\%} on GSM8K, \textbf{4.8\%} on MATH500, \textbf{42.2\%} on Countdown and \textbf{16.1\%} on Sudoku. 
\end{abstract}

\section{Introduction}
\label{sec:intro}
\begin{figure*}[t]
\centering  
    \begin{overpic}[width=1\linewidth]{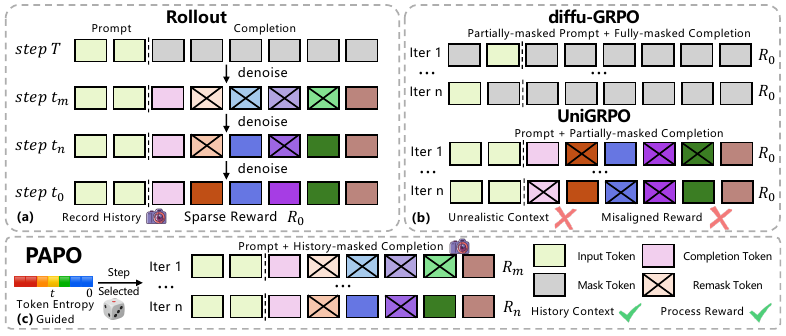}  
    \end{overpic} 
    \caption{\textbf{Comparison of RL Frameworks for dLLMs.} (a) Iterative denoising during the authentic rollout. (b) Existing methods perform policy updates by constructing artificial states from the final completion, introducing dual misalignment: the artificial context is unfaithful to the authentic generative process, and the sparse terminal reward ($R_0$) is indiscriminately assigned to all steps. (c) Our PAPO aligns the RL update with the authentic process by re-enacting historical states guided by entropy and assigning dense, step-aware process rewards ($R_m, R_n$).}
   
    \label{fig:motivation}
\end{figure*}
Diffusion Large Language Models (dLLMs) \citep{dream,gong2024scaling,llada,sahoo2024simple,sdar} have recently emerged as a significant paradigm shift from traditional auto-regressive (AR) generation, offering substantial inference speedups through their parallel generation process while achieving competitive performance. In contrast to the sequential token-by-token prediction of AR models, the core mechanism of dLLMs involves an iterative denoising and remasking loop that enables the parallel prediction and generation of multiple tokens at each step. However, while reinforcement learning (RL) has become a key technology for optimizing AR models~\cite{shao2024deepseekmath}, adapting these powerful interactive learning paradigms to the unique non-autoregressive architecture of dLLMs remains a critical and underexplored domain.

Recent efforts~\cite{sapo,tang2025wd1,zhao2025d1,yang2025mmada,wang2025time,gong2025diffucoder,zhan2025principled,wang2025revolutionizing,wang2025mro} have begun to adapt RL for dLLMs by integrating policy gradient methods~\cite{{ppo,dpo,shao2024deepseekmath}} into the non-autoregressive framework. 
Despite their remarkable progress, they are fundamentally constrained by a dual misalignment between the authentic generation trajectory and the gradient update process, as illustrated in Figure~\ref{fig:motivation}. 
First, existing methods exhibit a severe process-reward misalignment, where the sparse reward from the final outcome $R_{0}$ is indiscriminately assigned to decisions at any arbitrary intermediate step from $T$ to $t_0$, failing to provide step-specific credit assignment (Figure~\ref{fig:motivation} (b)). 
Second, they suffer from a lack of contextual state authenticity, policy updates are performed on ``artificial states'' constructed by randomly masking the final completion (Figure~\ref{fig:motivation} (b)), which lack faithfulness to the true intermediate contexts of the denoising process (Figure~\ref{fig:motivation} (a)). Moreover, their random step selection strategy leads to inefficient training by squandering updates on already settled decisions, at the expense of refining the critical steps where the model truly struggles.
This fundamental disconnect of the reward signal and policy update state from the authentic generative trajectory yields an inefficient and unstable optimization process, severely impeding the acquisition of structured reasoning abilities.

To address these fundamental misalignments, we introduce \textbf{P}rocess \textbf{A}ligned \textbf{P}olicy \textbf{O}ptimization (PAPO), a novel RL framework that holistically aligns the gradient update process with the authentic generative process of dLLMs. As shown in Figure~\ref{fig:motivation} (c), PAPO achieves this through two synergistic innovations. 
First, to tackle the process-reward misalignment, we introduce \textbf{S}tep-Aware \textbf{P}rocess \textbf{R}ewards (SPR). SPR provides a step-specific and dense credit assignment by calculating an immediate reward for each intermediate step based on a single denoising step prediction. This immediate reward is then integrated with the final outcome reward, forming a composite learning signal that provides accurate, step-specific credit.
Second, to resolve the challenge of unfaithful and inefficient state selection, we propose \textbf{E}ntropy-Guided \textbf{H}istorical \textbf{R}e-enactment (EHR). Departing from prior methods that rely on artificial states, EHR leverages authentic historical trajectories captured during the natural rollout process. To maximize learning efficiency, it re-enacts these trajectories for policy updates by employing an entropy-guided sampling strategy, which directs gradients toward moments of high model uncertainty, ensuring that learning is concentrated on the most critical stages of the generative process.

Extensive experiments on math reasoning (GSM8K, MATH500) and planning tasks (Countdown, Sudoku) demonstrate that \ourmethodabbr\ 
significantly outperforms the existing dLLM. Specifically, PAPO yields consistent improvements, with performance gains of up to $4.5\%$ on GSM8K and $4.8\%$ on MATH500. The advantages of our process aligned approach are even more pronounced on complex planning tasks, where PAPO achieves a staggering absolute improvement of $42.2\%$ on Countdown and $16.1\%$ on Sudoku. These results validate that by holistically aligning the RL updates with the model's generative process, PAPO provides a more effective and stable path to unlocking the advanced reasoning and planning capabilities of dLLMs.

The contributions of this paper are as follows:
\begin{itemize}
\item We identify two misalignments that form a core barrier to effective RL in dLLMs: process-reward misalignment and state-trajectory misalignment, both of which lead to inefficient and unstable policy updates.
\item We propose Process Aligned Policy Optimization (PAPO), a novel RL framework featuring Step-Aware Process Rewards (SPR) for process credit assignment and Entropy-Guided Historical Re-enactment (EHR) for history context state selection.
\item We demonstrate PAPO's superiority through extensive experiments, achieving state-of-the-art results with significant average gains of up to $4.7\%$ in math reasoning and $29.2\%$ in planning tasks over strong baselines.
\end{itemize}

\section{Related Work}
\label{sec:related work}
\textbf{Diffusion Large Language Models.}
Inspired by their success in continuous domains like image/video generation~\cite{ho2020denoising,song2020score,wan,shaovgpo,11573063,wang2024detectingdatasetabusefinetuning,wu2024improvedvideovaelatent}, diffusion models have recently been adapted for discrete text generation. While early efforts focused on continuous space~\cite{han2022ssd,li2022diffusion} or the probability simplex~\cite{avdeyev2023dirichlet,stark2024dirichlet}, the paradigm shifted with masked diffusion models~\cite{shi2024simplified,sahoo2024simple,nie2024scaling}, which rival GPT-2 level autoregressive models in perplexity~\cite{lou2023discrete}. This branched into two strategies: training massive models from scratch, exemplified by LLaDA~\cite{llada} that achieves performance competitive with LLaMA3-8B~\cite{dubey2024llama}, and converting pre-trained autoregressive models into diffusion generators, e.g. DiffuLLaMA~\cite{gong2024scaling} and Dream~\cite{dream}.
Concurrently, recent works have expanded their capabilities to include chain-of-thought reasoning~\cite{ye2024diffusion,ye2024beyond}, explored novel architectures like Block-Diffusion~\cite{arriola2025block,sdar} for structured generation, and extended the paradigm to multimodal models~\cite{yang2025mmada,Dai_2025,deng2025currreft,zhu2026doesvisualgenerationhelp,zhu2025llmknowsestimatingllmperceived, wei2026youtuvlunleashingvisualpotential,lu2025benchmarkinglargevisionlanguagemodels} like MMaDA.

\textbf{Reinforcement Learning for dLLMs.}
While RL has been widely successful in improving reasoning for autoregressive LLMs~\cite{shao2024deepseekmath,wu2026sequencemodelingalignmenttokenizer}, adapting it to dLLMs has recently attracted growing attention~\cite{rojas2026improvingreasoningdiffusionlanguage, wang2025mro,zhan2025principled,zhu2025enhancingreasoningdiffusionllms,zhao2025diffpotrainingdiffusionllms,ma2025consolidatingreinforcementlearningmultimodal,zhao2025inpaintingguidedpolicyoptimizationdiffusion}. Early efforts such as DRAKES~\cite{wang2025finetuningdiscretediffusionmodels} backpropagate rewards along the denoising trajectory but incur high computational cost and gradient instability. The dominant paradigm has thus shifted to direct policy optimization via GRPO~\cite{shao2024deepseekmath}, e.g., diffu-GRPO~\cite{zhao2025d1} and UniGRPO~\cite{yang2025mmada}. However, these methods suffer from two fundamental misalignments between RL updates and authentic trajectories.
First, process-reward misalignment: most methods assign sparse outcome rewards without process supervision, despite evidence that intermediate signals can usefully guide multi-step inference~\cite{chung2024diffusionposteriorsamplinggeneral,kan2025tacothinkanswerconsistencyoptimized,shao2025greatgeometryintentioncollaborativeinference,11475181}. TraceRL~\cite{wang2025revolutionizing} uses a value model with significant overhead. SAPO~\cite{sapo} requires costly Monte Carlo rollouts to estimate process rewards, limiting its efficiency.
Second, inefficient state selection: existing frameworks treat all trajectory steps equally. Although recent work~\cite{zhang2025right,chen2025seed,deng2026iiblpo} leverages entropy in RL, they apply it to advantage estimation rather than state selection for policy updates. We address both gaps via step-aware process rewards (SPR) for dense credit assignment and entropy-guided historical re-enactment (EHR) to focus learning on the most informative states.

\section{Preliminaries}
\textbf{Masked Diffusion Large Language Models.}
The paradigm of dLLMs~\cite{llada, dream} generates text through iterative denoising. The forward process corrupts a clean token sequence $x_0$ by stochastically replacing tokens with mask tokens over time $t \in [0, 1)$. The probability of each token remaining uncorrupted at timestep $t$ is determined by a noise schedule $\alpha_t$. The model $\pi_\theta$ learns to predict all masked tokens from $x_t$ by minimizing the NELBO. For masked dLLMs like LLaDA ~\cite{llada}, which employ a noise schedule $\alpha_t = 1-t$, this objective is reduced to a loss over masked tokens only:
\begin{equation}
\label{Eq:1}
\footnotesize
\boldsymbol{L}(\theta)=-\mathbb{E}_{t, x_{0}, x_{t} }\left[\frac{1}{t} \sum_{k=1}^{|x_t|} \mathbf{1}[x_t^k = \text{mask}] \log \pi_\theta(x_0^k | x_t) \right],
\end{equation}
where $\left | x_t \right | $ is the sequence length of $x$, and $x_0^k$ is the $k$-th token of $x_0$.

\textbf{Group Relative Policy Optimization.}
GRPO is a critic-free policy optimization method~\cite{shao2024deepseekmath} that derives advantages from group-based statistics. For a prompt $q$, it samples a group of $G$ responses $\{o_1, \dots, o_G\}$ and calculates an unnormalized group-relative advantage for each response $o_i$ based on its reward $R(o_i)$:
\begin{equation}
\label{Eq:2}
A_i = R(o_i) - \frac{1}{G} \sum_{j=1}^{G} R(o_j),
\end{equation}
where $R(\cdot)$ is the reward function. 

Adapting GRPO to dLLMs requires tractable sequence-level log-likelihoods $\log \pi_\theta(o|q)$, which autoregressive models obtain via the chain rule $\log \pi_\theta(o|q) = \sum_{k=1}^{|o|} \log \pi_\theta(o^k|q,o^{<k})$ but dLLMs cannot due to their non-causal, multi-step denoising process. Prevailing works~\cite{zhao2025d1,yang2025mmada} address this with a mean-field approximation, formulated as:
\begin{equation}
\label{Eq:3}
\log \pi_\theta(o|q) \approx \sum_{k=1}^{|o|} \log \pi_\theta(o^k|q),
\end{equation}
where each per-token marginal $\pi_\theta(o^k|q)$ can be computed via a single-pass estimate, replacing the need for explicit multi-step trajectory unrolling.

\begin{algorithm}[t]
\caption{\ourmethodabbr\ Training Process}
\label{alg:papo}
\footnotesize
\begin{algorithmic}[1]
\Require Reference model $\pi_{\text{ref}}$, prompts $\mathcal{D}$, completions $G$, number of inner updates $\mu$, diffusion steps $T$
\State Initialize policy $\pi_\theta \leftarrow \pi_{\text{ref}}$
\While{not converged}
    \State Set $\pi_{\text{old}} \leftarrow \pi_\theta$ and sample $q \sim \mathcal{D}$
    \State \Comment{\textit{// Rollout and Trajectory Recording}}
    \State Sample $\{o_i\}_{i=1}^G \sim \pi_{\text{old}}(\cdot | q)$, and cache $\{\tau_i\}_{i=1}^G$
    \State \Comment{\textit{// Step-Aware Process Rewards Computation}}
    \For{each rollout $o_i$ and trajectory $\tau_i$}
        \State Compute final sparse reward $R^f_{i} = R(o_i)$
        \State Compute step process reward $\{R_{i,t}^{p}\}_{t=1}^{T}$ by one-
        \State step decoding from each context $x_{i,t} \in \tau_i$
        \State Compute $R_{i,t}=R_i^f+R_{i,t}^p$ and $A_{i,t}$
    \EndFor
    \State \Comment{\textit{// Entropy-Guided Historical Re-enactment Update}}
    \For{gradient update iterations $n=1,\dots,\mu$}
        \State Compute entropies $\{H_{i,t}\}$ for $\tau_i$ with $\pi_{\text{old}}$
        \State Sample $t_n \sim p(t|i) \propto H_{i,t}^{\alpha}$ for $\tau_i$
        \State Reconstruct historical context $x_{i,t_n}$ from $\tau_i$
        \State Estimate log-probabilities under $\pi_\theta,\pi_{\text{old}},\pi_{\text{ref}}$
        \State Compute PAPO objective $\boldsymbol{L}_{papo}$ and update $\pi_\theta$
    \EndFor
\EndWhile
\State \Return $\pi_\theta$
\end{algorithmic}
\end{algorithm}

\section{\ourmethod}
Current RL frameworks for dLLMs suffer from two fundamental misalignments: sparse, outcome-based rewards that fail to credit the reasoning process, and inefficient, unfaithful state selection for policy updates. To address these shortcomings, we introduce Process Aligned Policy Optimization (PAPO), which aligns learning with the dLLM's generative trajectory through Step-Aware Process Rewards (SPR) (Section~\ref{3.2.1}) and Entropy-Guided Historical Re-enactment (EHR) (Section~\ref{3.2.2}). The complete algorithm is detailed in Algorithm \ref{alg:papo}.

\begin{figure}[t]
\centering  
    \begin{overpic}[width=1\linewidth]{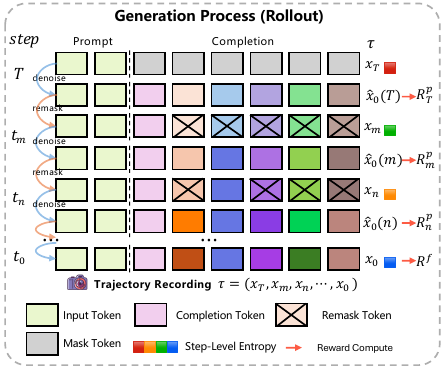} 
    \end{overpic} 
    \vspace{-6mm}
    \caption{\textbf{Detailed Rollout Process of PAPO: Trajectory Recording and Process Reward Computation.} This figure illustrates how PAPO captures the complete generative trajectory $\tau = (x_{T}, \dots, x_{t}, \dots, x_{0})$ while concurrently computing Step-Aware Process Rewards (SPR) $R^p_t$ based on the one-step denoised prediction $\hat{x}_0{(t)}$ from each intermediate state $x_t$.} 
    \label{fig:method_rollout}
\end{figure}

\subsection{Step-Aware Process Rewards}
\label{3.2.1}
A core limitation of existing RL methods for dLLMs is the  process-reward mismatch: a single, sparse reward is assigned based on the final outcome, providing no supervision on the quality of the intermediate reasoning path and making it insufficient for guiding the complex, multi-step generation process of dLLMs. To resolve it, we introduce Step-Aware Process Rewards (SPR) to generate dense, fine-grained rewards evaluating each intermediate step in the denoising trajectory.

Given a prompt $q$, the dLLM iteratively refines from a fully masked state $x_T$. At each step $t$, the model predicts all masked tokens from $x_t$ to produce a one-step denoised prediction, then remasks remaining tokens to form the next state $x_{t-1}$, continuing until the final completion $o_i$. In conventional RL, policy then guides updates for every intermediate step using a sparse reward $R^f_{i} = R(o_i)$.

Our SPR instead records the full generative trajectory $\tau_i = (x_{T}, \dots, x_{t}, \dots, x_{0})$ for each rollout $i$, as illustrated in Figure~\ref{fig:method_rollout}.
At each step $t$, we form the one-step denoised prediction $\hat{x}_0(t)$, which is the complete solution estimate from context $x_t$ before remasking, and evaluate the step-specific process reward $R_{i,t}^{p}$ with the reward function:
\begin{equation}
\label{Eq:4}
R_{i,t}^{p} = R(\hat{x}_0(t)), \hat{x}_0(t) \sim \pi_{\text{old}}(\cdot | q, x_t),
\end{equation}
where $\hat{x}_0(t)$ is a one-step denoised prediction from $x_t$ using the old policy $\pi_{\text{old}}$. 

To ground this immediate feedback with the long-term objective, dense signal is combined with the final outcome reward to form a comprehensive, step-specific total reward:
\begin{equation}
\label{Eq:5}
R_{i,t} = R^f_{i} + R_{i,t}^{p}.
\end{equation}
From this composite reward, a more accurate, step-aware advantage $A_{i,t}$ is computed using Eq.~\ref{Eq:2} and cached alongside $\tau_i$ for subsequent policy updates.

\subsection{Entropy-Guided Historical Re-enactment}
\label{3.2.2}
The second misalignment is inefficient and unfaithful state selection for policy updates. As shown in Figure~\ref{fig:motivation} (b), prevalent methods such as diffu-GRPO~\cite{zhao2025d1} and UniGRPO~\cite{yang2025mmada} construct artificial contexts that do not reflect the model's authentic generation process. The diffu-GRPO performs its updates by conditioning on a synthetic state composed of a randomly masked prompt $q'$ and a fully masked completion $\pi_{\theta}(o^k|q'\oplus mask\dots \oplus mask)$, which is a zero-information context the model never encounters post-initialization. While UniGRPO forms its context by applying a random mask ratio $p_i \in [0,1]$ directly to the final completion $o_i$. This creates an unrealistic context that is unlikely to correspond to actual state from the generative trajectory. In both cases, a fundamental discrepancy arises between the synthetic contexts used for learning and the authentic states the policy actually encounters during inference, resulting in suboptimal gradient updates.

To resolve this state-selection discrepancy, we introduce Entropy-Guided Historical Re-enactment (EHR) to select authentic and efficient states for policy updates. EHR uses the cached trajectories $\tau_i = (x_{T}, \dots, x_{t}, \dots, x_{0})$ and their step-aware advantages $\{A_{i,t}\}$ to perform policy updates directly on authentic intermediate states $x_{i,t}$, eliminating the train-inference distribution mismatch.

As naive uniform sampling becomes inefficient by repeatedly revisiting well-learned, low-entropy steps, EHR instead prioritizes high-entropy states where policy uncertainty is greatest, maximizing the utility of each gradient update.
Specifically, EHR computes the step-level entropy $H_{i,t}$ for each timestep $t$ as the average token-level entropy over masked positions $k \in \mathcal{M}_{i,t}$:
\begin{equation}
\label{eq:entropy}
H_{i,t} = \frac{1}{|\mathcal{M}_{i,t}|} \sum_{k \in \mathcal{M}_{i,t}} \left( - \sum_{v \in \mathcal{V}} p_k(v) \log p_k(v) \right),
\end{equation}
where $\mathcal{M}_{i,t}$ is the set of masked tokens in state $x_{i,t}$, and $p_k(v) = \pi_{\text{old}}(v | q, x_{i,t}, k)$ denotes the probability assigned by the old policy $\pi_{\text{old}}$ to a specific token $v$ from the vocabulary $\mathcal{V}$ for the masked position $k$. We construct a timestep sampling distribution where the probability of selecting step $t_n$ is proportional to its entropy:
\begin{equation}
p(t_n | i) \propto (H_{i,t_n})^\alpha,
\end{equation}
where $\alpha$ controls sharpness of the distribution: $\alpha=0$ yields uniform sampling, and larger $\alpha$ increasingly favors high-uncertainty steps.

At each update iteration, we sample $t_n \sim p(t|i)$ and reconstruct authentic context $x_{i,t_n}$ from $\tau_i$. Integrating SPR-derived advantages with EHR state selection, the complete PAPO objective is:
\begin{equation} 
\label{eq:papo_objective}
\resizebox{1.0\columnwidth}{!}{ 
$\begin{gathered}
\mathcal{L}_{\text{PAPO}}(\theta) = \mathbb{E}_{\substack{q \sim \mathcal{D},\{\tau_i\} \sim \pi_{\text{old}}(\cdot|q) \\ t_n \sim p(\cdot|i)}} \Bigg[  \bigg( \frac{1}{G} \sum_{i=1}^G \frac{1}{|\mathcal{M}_{i,t_n}|} \sum_{k \in \mathcal{M}_{i,t_n}} \min \Big( r_{i,t_n}^k(\theta) A_{i,t_n}, \\
\text{clip}(r_{i,t_n}^k(\theta), 1-\epsilon, 1+\epsilon) A_{i,t_n} \Big) \bigg) - \beta D_{KL}(\pi_\theta || \pi_{\text{ref}})\Bigg],
\end{gathered}$
}
\end{equation}
where $A_{i,t_n}$ is the cached step-aware advantage and $r_{i,t_n}^k(\theta) = \frac{\pi_\theta(o_i^k | q, x_{i,t_n})}{\pi_{\text{old}}(o_i^k | q, x_{i,t_n})}$ is the importance ratio. $\epsilon$ and $\beta$ control clipping and KL regularization. 

\section{Experiments}

\definecolor{greenx}{HTML}{00CD00}
\begin{table*}[t]
\vspace{-4mm}
\begin{center}
\resizebox{\linewidth}{!}{
\centering
\footnotesize
\renewcommand{\arraystretch}{1.2} 
\renewcommand{\tabcolsep}{6.5 pt}
\begin{tabular}{l ccc ccc ccc ccc}
\toprule
 & \multicolumn{3}{c}{\textbf{GSM8K}} & \multicolumn{3}{c}{\textbf{MATH500}} & \multicolumn{3}{c}{\textbf{Countdown}} & \multicolumn{3}{c}{\textbf{Sudoku}} \\
\cmidrule(lr){2-4} \cmidrule(lr){5-7} \cmidrule(lr){8-10} \cmidrule(lr){11-13}
\textbf{Model / Seq Len} & \textbf{128} & \textbf{256} & \textbf{512} & \textbf{128} & \textbf{256} & \textbf{512} & \textbf{128} & \textbf{256} & \textbf{512} & \textbf{128} & \textbf{256} & \textbf{512} \\
\midrule
LLaDA-8B-Instruct~\cite{llada} & 68.7 & 76.7 & 78.2 & 26.0 & 32.4 & 36.2 & 20.7 & 19.5 & 16.0 & 11.7 & 6.7 & 5.5 \\
diffu-GRPO~\cite{zhao2025d1} & 72.6 & 79.8 & 81.9 & 33.2 & 37.2 & 39.2 & 33.2 & 31.3 & 37.1 & 18.4 & 12.9 & 11.0 \\
TSE-Vote~\cite{wang2025time} & 70.1 & 78.7 & 78.9 & 28.4 & 35.6 & 36.2 & 25.0 & 23.4 & 16.4 & $\times$ & $\times$ & $\times$ \\
WINO~\cite{wino} & - & 75.8 & - & - & 34.2 & - & - & 33.2 & - & - & 15.2 & - \\
UniGRPO$^{\dagger}$~\cite{yang2025mmada} & 73.2 & 79.8 & 79.4 & 33.0 & 37.0 & 39.8 & 51.6 & 53.9 & 43.8 & 23.9 & 19.2 & 14.1 \\
SAPO~\cite{sapo} & 72.9 & 82.2 & \underline{82.4} & 32.0 & \textbf{40.0} & 38.4 & 51.6 & 52.0 & 56.3 & 22.4 & 20.3 & 16.1 \\
\midrule
SFT & 66.5 & 78.8 & 81.1 & 26.2 & 32.6 & 34.8 & 20.3 & 14.5 & 23.8 & 16.5 & 8.5 & 4.6 \\
SFT + diffu-GRPO~\cite{zhao2025d1} & 73.2 & 81.1 & 82.1 & \textbf{33.8} & 38.6 & 40.2 & 34.8 & 32.0 & 42.2 & 22.1 & 16.7 & 9.5 \\
SFT + TSE-Reward~\cite{wang2025time} & 72.1 & 80.0 & \textbf{83.0} & 31.2 & 35.4 & \textbf{41.4} & 41.5 & 42.6 & 54.7 & $\times$ & $\times$ & $\times$ \\
\midrule
\multirow{2}{*}{\textbf{Ours}} & \textbf{73.8} & \textbf{82.4} & 80.8 & \underline{33.4} & 35.6 & \underline{40.0} & \textbf{52.0} & \textbf{65.6} & \textbf{65.2} & \textbf{27.2} & \textbf{25.0} & \textbf{20.2} \\
& \textcolor{greenx}{\textbf{+5.1}} & \textcolor{greenx}{\textbf{+5.7}} & \textcolor{greenx}{\textbf{+2.6}} & \textcolor{greenx}{\textbf{+7.4}} & \textcolor{greenx}{\textbf{+3.2}} & \textcolor{greenx}{\textbf{+3.8}} & \textcolor{greenx}{\textbf{+31.3}} & \textcolor{greenx}{\textbf{+46.1}} & \textcolor{greenx}{\textbf{+49.2}} & \textcolor{greenx}{\textbf{+15.5}} & \textcolor{greenx}{\textbf{+18.3}} & \textcolor{greenx}{\textbf{+14.7}} \\
\bottomrule
\end{tabular}
}
\caption{\textbf{Performance Comparison on Four Benchmarks.} \textbf{Bold} numbers indicate the top performance in each category.  \underline{Underlined} numbers denote the best performance among methods without additional Supervised Fine-Tuning (SFT). \textcolor{greenx}{Green values} show the absolute improvement of our method over the LLaDA-8B-Instruct baseline. ``–'' denotes unreported results, ``×'' denotes unsupported tasks, and ${\dagger}$ indicates our own re-implementation. Without additional SFT, our PAPO demonstrates superior
performance.}
\label{table:all}
\end{center}
\end{table*}
\begin{figure*}[t]
\centering  
    \begin{overpic}[width=1\linewidth]{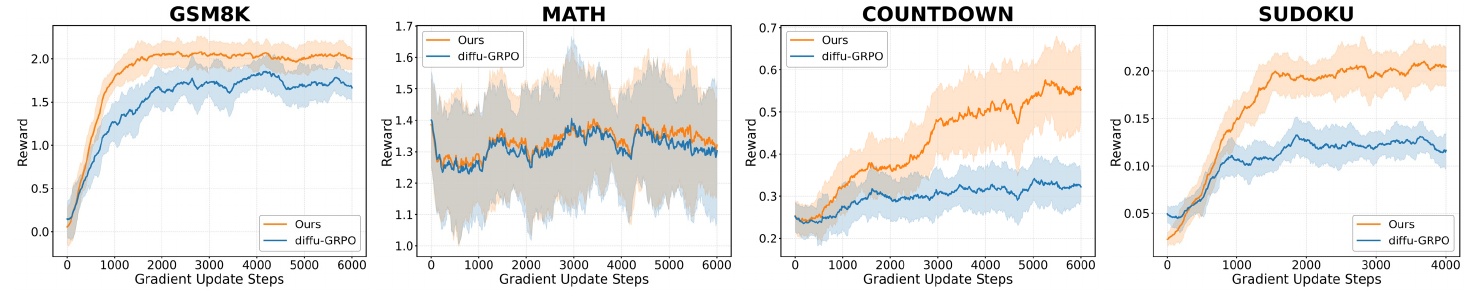} 
    \end{overpic} 
    \vspace{-4mm}
    \caption{\textbf{Reward Curves during RL Training.} Comparison of our method (PAPO) and diffu-GRPO across four benchmarks. PAPO consistently achieves higher reward trajectories, demonstrating superior sample efficiency.} 
    \label{fig:result}
    \vspace{-2mm}
\end{figure*}

\begin{table*}[!t]
\begin{center}
\tiny
\resizebox{\linewidth}{!}{
\renewcommand{\arraystretch}{1} 
\renewcommand{\tabcolsep}{6.5 pt}
\begin{tabular}{l ccc ccc ccc ccc}
\toprule
 & \multicolumn{3}{c}{\textbf{GSM8K}} & \multicolumn{3}{c}{\textbf{MATH500}} & \multicolumn{3}{c}{\textbf{Countdown}} & \multicolumn{3}{c}{\textbf{Sudoku}} \\
\cmidrule(lr){2-4} \cmidrule(lr){5-7} \cmidrule(lr){8-10} \cmidrule(lr){11-13}
\textbf{Model / Seq Len} & \textbf{128} & \textbf{256} & \textbf{512} & \textbf{128} & \textbf{256} & \textbf{512} & \textbf{128} & \textbf{256} & \textbf{512} & \textbf{128} & \textbf{256} & \textbf{512} \\
\midrule
LLaDA-8B-Instruct & 68.7 & 76.7 & 78.2 & 26.0 & 32.4 & 36.2 & 20.7 & 19.5 & 16.0 & 11.7 & 6.7 & 5.5 \\
\quad + SPR & 71.6 & 80.2 & 79.3 & 32.6 & 37.2 & 37.8 & 47.7 & 55.5 & 55.1 & 24.8 & 23.3 & 17.2 \\
\quad + HR & 73.8 & 80.1 & 80.2 & 32.2 & 36.8 & 36.4 & 51.9 & 56.6 & 55.5 & 25.3 & 22.6 & 18.7  \\
\quad + HR + SPR& \textbf{74.0} & 81.3 & 79.8 & 32.2 & \textbf{37.6} & 38.0 & 51.6 & 60.6 & 58.2 & 25.8 & 23.6 & 18.9 \\
\quad + EHR & 73.2 & 81.1 & 80.5 & 32.8 & 36.4 & 37.0 & 51.2 & 57.5 & 56.3 & 25.7 & 23.4 & 17.3 \\
\quad + EHR + SPR (PAPO) & 73.8 & \textbf{82.4} & \textbf{80.8} & \textbf{33.4} & 35.6 & \textbf{40.0} & \textbf{52.0} & \textbf{65.6} & \textbf{65.2} & \textbf{27.2} & \textbf{25.0} & \textbf{20.2}\\
\bottomrule
\end{tabular}
}

\caption{\textbf{Ablation Study} on the contributions of Step-Aware Process Rewards (SPR) and Entropy-Guided Historical Re-enactment (EHR), where HR denotes Historical Re-enactment without entropy guidance.}
\label{table:ablation}
\end{center}
\vspace{-4mm} 
\end{table*}

\begin{figure*}[t]
\centering  
    \begin{overpic}[width=1\linewidth]{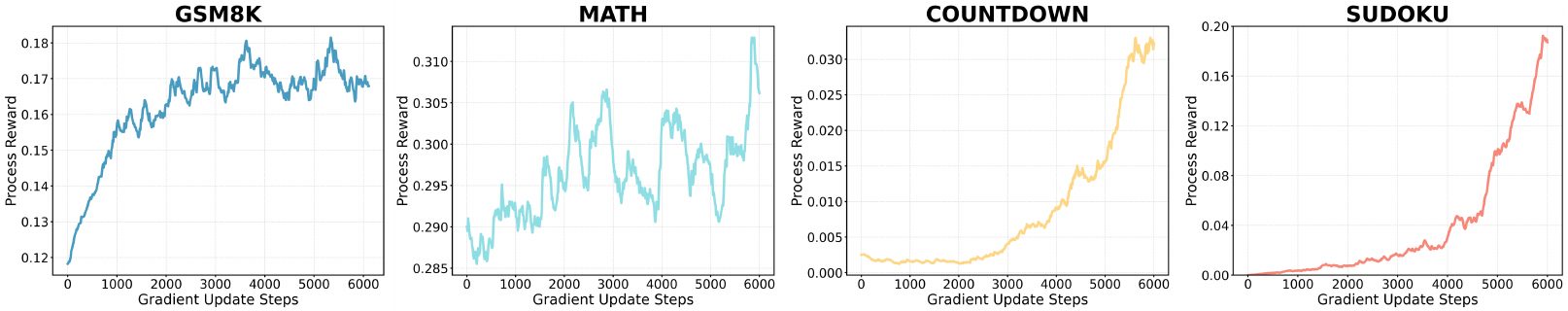} 
    \end{overpic} 
    \vspace{-6mm}
    \caption{\textbf{Training Dynamics of Step-Aware Process Rewards.} The consistent upward trend of the process reward across all benchmarks confirms that the policy is learning to generate higher-quality intermediate reasoning steps.} 
    \label{fig:pr}
\end{figure*}

\subsection{Experimental Setup} 
\textbf{Tasks.}
We evaluate our method on four reasoning tasks: (1) GSM8K~\cite{gsm}, a grade school math dataset requiring multi-step logical inference; (2) MATH500~\cite{math}, a curated set of 500 competition level high-school math problems; (3) Countdown, a combinatorial arithmetic game requiring target expression generation; (4) 4x4 Sudoku, a planning task demanding constraint satisfaction and systematic elimination.

\textbf{Compared Baselines.}
 We compare PAPO with several RL methods for dLLMs. Our primary baselines include d1~\cite{zhao2025d1}, the first work to apply RL to dLLMs and diffu-GRPO applied directly to the base model. For a comprehensive comparison, we additionally evaluate against recent SOTA RL methods: UniGRPO~\cite{yang2025mmada}, WINO~\cite{wino}, TSE~\cite{wang2025time}, SAPO~\cite{sapo} as well as models further fine-tuned on the reasoning dataset s1k~\cite{s1}. 

\textbf{Training Details.}
All experiments are based on the LLaDA-8B-Instruct model~\cite{llada}, using official splits for GSM8K and MATH, and the same synthetic datasets as d1~\cite{zhao2025d1} for Countdown and Sudoku. Unlike d1, PAPO applies RL directly to the base model without supervised fine-tuning (SFT). During RL training, the sequence length for online trajectory generation is fixed at 256 tokens, while evaluation spans 128, 256, and 512 tokens to assess generalization. All experiments are conducted on 8 NVIDIA A100 GPUs. Please refer to Appendix~\ref{A} for more details.


\subsection{Main Results} 
\textbf{Overall Performance.} As detailed in Table~\ref{table:all}, PAPO demonstrates superior performance across four challenging reasoning benchmarks. 
Without any SFT, PAPO achieves an average absolute improvement of 16.9\% over LLaDA-8B-Instruct, establishing state-of-the-art results among direct RL methods on the majority of tasks. The gains are especially pronounced on planning tasks, reaching up to 42.2\% on Countdown and 16.1\% on Sudoku, where structured, multi-step credit assignment proves most critical. 
These results confirm that PAPO unlocks advanced reasoning by holistically aligning the learning signal with the generative process, an alignment powered by the strategic integration of process-aware rewards and authentic state selection.

\textbf{Training Dynamics.} As shown in Figure~\ref{fig:result}, PAPO exhibits markedly superior training dynamics compared to the diffu-GRPO baseline, achieving a consistently higher reward trajectory with a steeper ascent and lower volatility. This rapid and stable improvement reflects the superior sample efficiency of PAPO, indicating that our method more effectively leverages the learning signal to accelerate policy optimization. 
We attribute these gains to the synergy between EHR, which provides more informative training states, and SPR, which delivers granular, process-aware feedback in contrast to sparse, terminal rewards.

\begin{table}[t]
\centering
\footnotesize
\renewcommand{\arraystretch}{1.}
\renewcommand{\tabcolsep}{4pt}
\resizebox{\linewidth}{!}{
\begin{tabular}{ccccc}
\toprule
 & \textbf{1-step} & \textbf{4-step} & \textbf{8-step} & \textbf{16-step} \\
\midrule
Seq Len=128 & 73.8 & 74.2\textcolor{darkpink}{$\diamond$0.4\%} & 74.7\textcolor{darkpink}{$\diamond$0.9\%} & \textbf{75.4}\textcolor{darkpink}{$\diamond$1.6\%} \\
GPU Hours & \textbf{177} & 187\textcolor{darkpink}{$\diamond$5.6\%} & 200\textcolor{darkpink}{$\diamond$13.0\%} & 255\textcolor{darkpink}{$\diamond$44\%} \\
\bottomrule
\end{tabular}
}
\caption{\textbf{Multi-step SPR Comparison on GSM8K.} Increasing the lookahead steps yields diminishing accuracy gains at prohibitive computational cost. $\textcolor{darkpink}{\diamond}$ denotes the improvement over 1-step.}
\label{table:multi-step}
\end{table}

\subsection{Ablation Study}
We conduct a thorough ablation study to examine the contribution of each component, as shown in Table~\ref{table:ablation}. First, removing SPR and reverting to sparse, terminal-only rewards leads to a substantial performance drop, underscoring the necessity of dense, process-aware feedback for effective credit assignment. Second, replacing EHR with a simpler Historical Re-enactment (HR) that samples states uniformly also yields a noticeable performance decline, demonstrating that training on authentic states alone is insufficient: prioritizing high-entropy states where the policy is most uncertain is critical for maximizing learning efficiency. The full PAPO model achieves the best overall performance, confirming that the strength of our approach lies in directing granular, process-aware rewards toward the most informative historical states.

We further analyze why SPR and EHR yield larger gains when combined with state alignment.
First, SPR computes process rewards on authentic rollout states $x_t \in \tau_i$.
In the diffu-GRPO setting, policy updates are performed on randomly masked states that do not correspond to the rollout trajectory. Applying dense process rewards estimated at authentic states to guide optimization at misaligned states therefore fails to accurately credit the corresponding actions.
Once optimization states are aligned with the rollout trajectory via EHR, these step-specific rewards regain their intended meaning and enable more precise credit assignment.
Second, entropy-guided selection assigns higher sampling probability to high-uncertainty steps along the authentic denoising trajectory, which presupposes a structured trajectory with meaningful entropy variation across steps.
Under Historical Re-enactment (HR), authentic states naturally provide this structure, allowing entropy guidance to concentrate learning on the most critical decision points.
Together, these observations explain the complementary gains of SPR and entropy guidance when paired with state-aligned optimization.

\begin{figure}[t]
\centering
\begin{subfigure}{0.48\columnwidth}
  \centering
  \includegraphics[width=\linewidth]{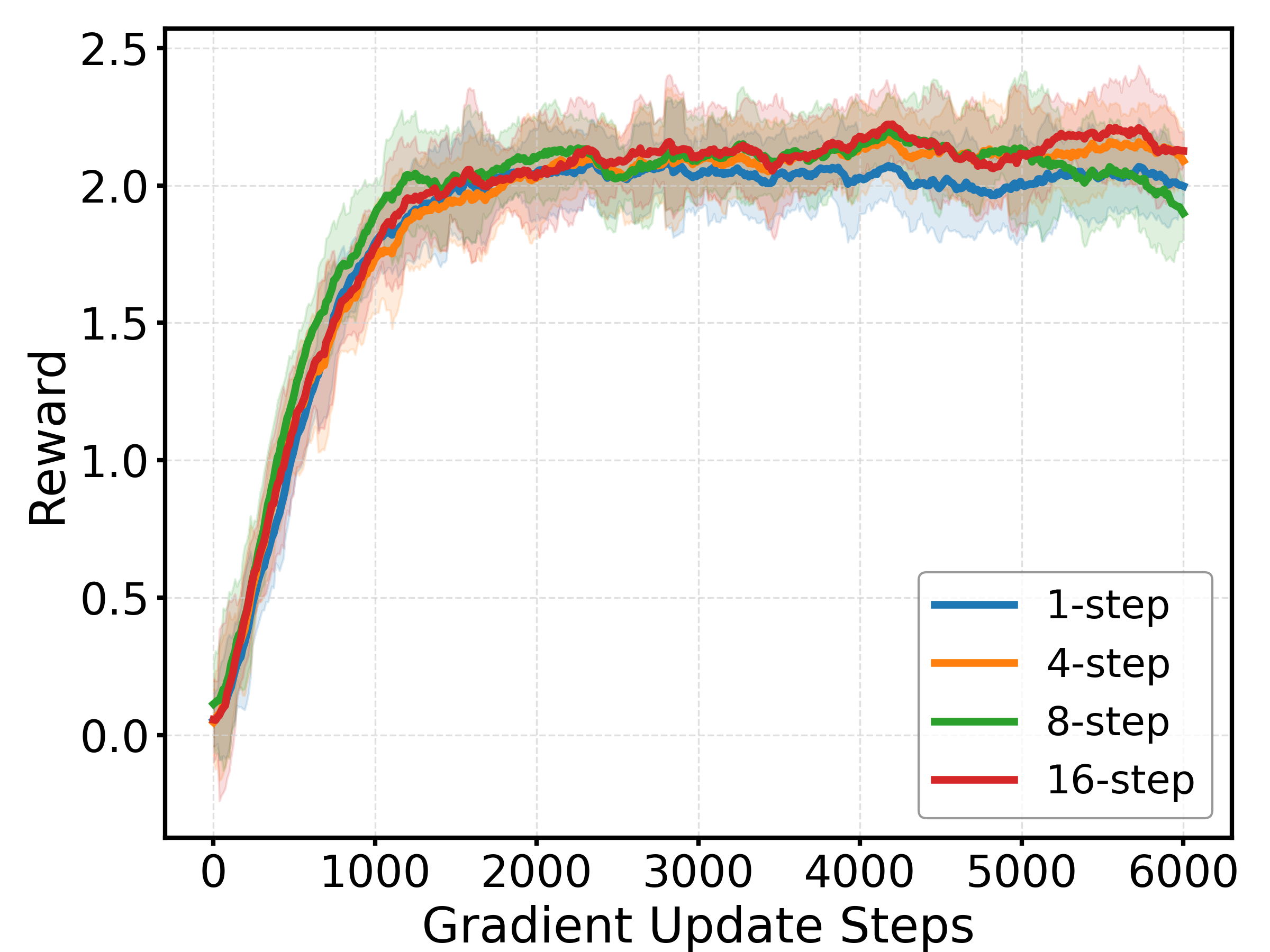}
  \caption{Multi-step comparison}
  \label{fig:multi_step}
\end{subfigure}
\hfill
\begin{subfigure}{0.48\columnwidth}
  \centering
  \includegraphics[width=\linewidth]{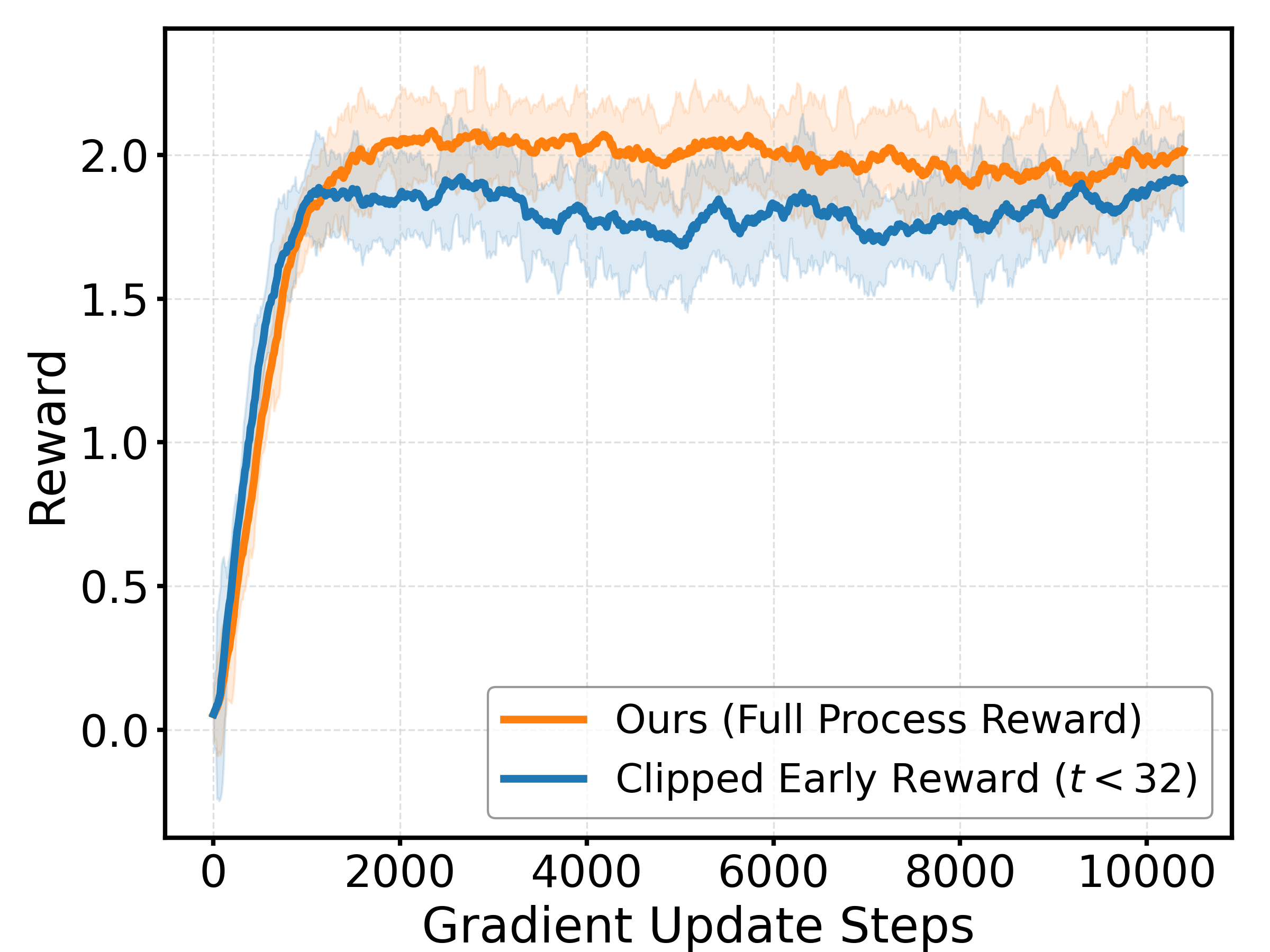}
  \caption{Early reward clipping}
  \label{fig:clip_reward}
\end{subfigure}

\caption{\textbf{SPR Fidelity-Efficiency Analysis.} (a) Multi-step SPR training dynamics on GSM8K. (b) Clipping early-stage rewards ($t < 32$) degrades performance, confirming the importance of early supervision.}
\label{fig:spr_analysis}
\end{figure}

\begin{figure*}[t]
\centering  
    \begin{overpic}[width=1\linewidth]{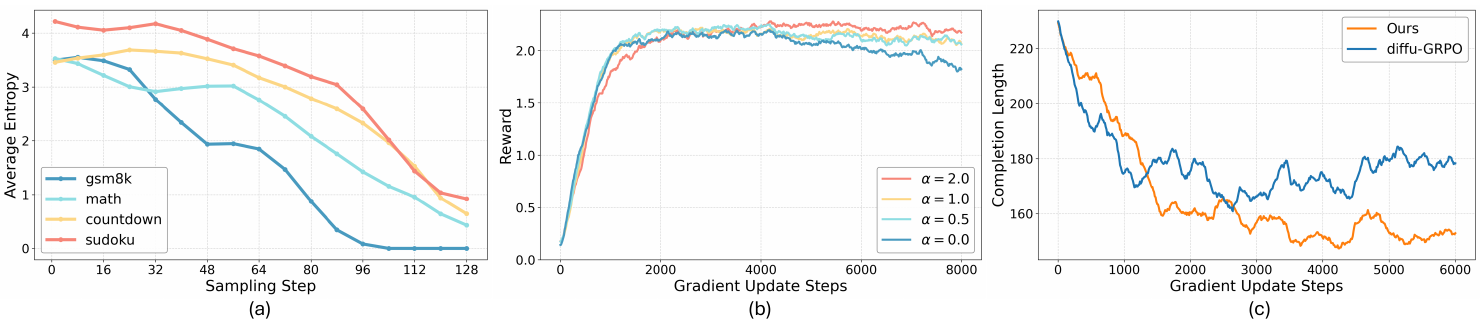} 
    \end{overpic} 
    \vspace{-6mm}
    \caption{\textbf{Analysis of Entropy-Guided Historical Re-enactment.} (a) Average token-level entropy exhibits a clear downward trend as the number of sampling steps increases. (b) The impact of the entropy-weighting hyperparameter $\alpha$  on the reward trajectory. (c) Completion length dynamics of PAPO and diffu-GRPO on the GSM8K benchmark.} 
    \label{fig:entropy}
\end{figure*}

\begin{table*}[t]
\begin{center}
\tiny
\resizebox{\linewidth}{!}{
\renewcommand{\arraystretch}{1}
\renewcommand{\tabcolsep}{6.5pt}
\begin{tabular}{l ccc ccc ccc ccc}
\toprule
& \multicolumn{3}{c}{\textbf{GSM8K}} & \multicolumn{3}{c}{\textbf{MATH500}} & \multicolumn{3}{c}{\textbf{Countdown}} & \multicolumn{3}{c}{\textbf{Sudoku}} \\
\cmidrule(lr){2-4} \cmidrule(lr){5-7} \cmidrule(lr){8-10} \cmidrule(lr){11-13}
\textbf{Model / Seq Len} & \textbf{128} & \textbf{256} & \textbf{512} & \textbf{128} & \textbf{256} & \textbf{512} & \textbf{128} & \textbf{256} & \textbf{512} & \textbf{128} & \textbf{256} & \textbf{512} \\
\midrule
LLaDA-1.5 & 69.8 & 79.4 & 81.1 & 29.0 & 32.4 & 35.4 & 21.5 & 21.1 & 20.7 & 12.4 & 8.7 & 7.3 \\
diffu-GRPO & 73.0 & 78.9 & \textbf{83.1} & 29.8 & 36.2 & \textbf{40.2} & 38.7 & 29.7 & 39.1 & 23.5 & 18.3 & 13.1 \\
\rowcolor{mygray}
\textbf{Ours} & \textbf{75.9} & \textbf{81.7} & 81.8 & \textbf{31.2} & \textbf{36.6} & 39.4 & \textbf{40.6} & \textbf{68.8} & \textbf{66.0} & \textbf{44.8} & \textbf{26.8} & \textbf{24.2} \\
\bottomrule
\end{tabular}
}
\end{center}
\caption{\textbf{Generalization to LLaDA-1.5.} PAPO consistently outperforms diffu-GRPO across all benchmarks, with the largest gains on planning tasks.}
\label{table:llada1_5}
\end{table*}

\section{Analysis}
\subsection{Effect of Step-Aware Process Rewards}
\textbf{Process reward drives structured reasoning.}
To validate the effectiveness of SPR, we track the average process reward during training (Figure~\ref{fig:pr}). The consistent upward trend confirms that the policy progressively learns to produce higher-quality intermediate reasoning steps, rather than merely arriving at correct final answers by chance. This dense, step-level supervision directly addresses the ``unstructured refinement'' problem by incentivizing coherent, step-by-step reasoning paths.

\textbf{Fidelity-Efficiency trade-off favors one-step SPR.}
A natural question is whether one-step denoised predictions faithfully reflect generation quality. We take GSM8K as a representative case to examine the fidelity-efficiency trade-off. As shown in Table~\ref{table:multi-step} and Figure~\ref{fig:spr_analysis}\,(a), increasing the lookahead from 1-step to 16-step yields only marginal accuracy gains (+1.6\%)  while incurring substantial overhead (+44\% GPU hours), confirming that 1-step SPR achieves the optimal fidelity-efficiency trade-off. We further examine the role of early-stage process rewards. As shown in Figure~\ref{fig:spr_analysis}\,(b), clipping
rewards for steps $t < 32$ leads to noticeable performance
degradation, confirming that early-stage process rewards
are critical for stable convergence.

\subsection{Effect of Entropy-Guided Historical Re-enactment}
\textbf{Entropy-guided selection improves learning efficiency.}
The generative process exhibits a non-uniform entropy distribution across the denoising trajectory (Figure~\ref{fig:entropy}\,(a)), with entropy decreasing as denoising progresses. Uniform timestep selection is thus inefficient, over-sampling low-entropy states where the policy is confident while under-sampling high-entropy states with the greatest learning potential. EHR addresses this by probabilistically prioritizing updates on high-entropy states.

\textbf{Impact of hyperparameter $\alpha$ in EHR.}
We further validate this design through an analysis of $\alpha$ (Figure~\ref{fig:entropy}\,(b)). Without entropy guidance ($\alpha=0$), training yields the lowest reward and later-stage instability, as uniform sampling keeps updating well-learned states. With entropy weighting ($\alpha>0$), reward and convergence stability both improve, confirming that prioritizing high-entropy states benefits learning efficiency and training stability.

\textbf{Token Efficiency.}
The efficiency gain also manifests in the generated outputs. As observed on the GSM8K benchmark (Figure~\ref{fig:entropy}\,(c)), PAPO converges to shorter completions than diffu-GRPO while achieving higher task performance, indicating improved token efficiency. This suggests that process aligned optimization steers the policy toward more decisive reasoning trajectories, reducing redundant tokens without sacrificing quality.

\subsection{Cross-Domain Generalization}
\textbf{Generalization across Backbones.}
We apply PAPO to the LLaDA-1.5~\cite{zhu2025llada15variancereducedpreference} model to assess cross-backbone transfer (Table~\ref{table:llada1_5}). PAPO consistently outperforms diffu-GRPO on all benchmarks, with pronounced gains on planning tasks, demonstrating that the process aligned optimization transfers effectively across model scales.

\begin{table}[t]
\begin{center}
\tiny
\resizebox{\linewidth}{!}{
\renewcommand{\arraystretch}{1}
\renewcommand{\tabcolsep}{6pt}
\begin{tabular}{l ccc ccc}
\toprule
& \multicolumn{3}{c}{\textbf{HumanEval}} & \multicolumn{3}{c}{\textbf{MBPP}} \\
\cmidrule(lr){2-4} \cmidrule(lr){5-7}
\textbf{Model / Seq Len} & \textbf{128} & \textbf{256} & \textbf{512} & \textbf{128} & \textbf{256} & \textbf{512} \\
\midrule
LLaDA + SFT & 21.3 & 32.3 & 32.9 & 40.1 & 39.7 & 41.2 \\
diffu-GRPO & 31.1 & 32.9 & 37.8 & 40.5 & \textbf{44.7} & 42.8 \\
\rowcolor{mygray}
\textbf{Ours} & \textbf{31.7} & \textbf{34.8} & \textbf{40.2} & \textbf{42.1} & 43.2 & \textbf{47.1} \\
\bottomrule
\end{tabular}
}
\caption{\textbf{Generalization to Code Task.} PAPO achieves competitive or superior performance on HumanEval and MBPP with execution-based rewards.}
\label{table:code}
\end{center}
\end{table}

\textbf{Generalization across Task Domains.}
We further evaluate PAPO on code generation (Table~\ref{table:code}), a domain involving more complex, execution-based reward functions. We train a model on the KodCode-Light-RL-10K dataset~\cite{xu2025kodcodediversechallengingverifiable}. PAPO achieves consistently competitive or superior results on both HumanEval and MBPP benchmarks, confirming that our framework is not restricted to mathematical and planning tasks.

\textbf{Generalization to More Challenging Tasks.}
Beyond transferring across backbones and task domains, we investigate whether PAPO scales to planning problems with substantially larger search spaces and longer reasoning horizons. To this end, we extend our evaluation from 4$\times$4 to 9$\times$9 Sudoku. We randomly sample 500 instances from Puzzle Bench for evaluation and use the remaining instances for RL training.

As shown in Table~\ref{table:sudoku9_llada}, LLaDA-8B-Instruct achieves below 1\% accuracy across all sequence lengths, highlighting the difficulty of this setting. Nevertheless, PAPO improves its average accuracy from 0.60\% to 7.32\%, an absolute gain of 6.72 points, and consistently outperforms diffu-GRPO.

Because RL performance is inherently constrained by the capability of the base model, we additionally evaluate PAPO on SDAR~\cite{sdar}, a stronger block-diffusion model converted from a well-trained autoregressive model. SDAR generates outputs blockwise and therefore does not use a fixed sequence length. Its accuracy increases from 2.65\% to 19.99\% after PAPO training, compared with 13.32\% for diffu-GRPO. These results demonstrate that PAPO remains effective on substantially harder planning tasks and generalizes across diffusion architectures.

\begin{table}[t]
\centering
\tiny
\resizebox{\linewidth}{!}{
\renewcommand{\arraystretch}{1}
\renewcommand{\tabcolsep}{6pt}
\begin{tabular}{lcccc}
\toprule
Model / Seq. Len. & 128 & 256 & 512 & Avg. \\
\midrule
LLaDA-8B-Instruct & 0.60 & 0.99 & 0.21 & 0.60 \\
diffu-GRPO$^\dagger$ & 3.48 & 4.52 & 2.67 & 3.56 \\
\rowcolor{mygray}
\textbf{Ours}& \textbf{7.15} & \textbf{8.06} & \textbf{6.76} & \textbf{7.32} \\
\bottomrule
\end{tabular}
}
\caption{
\textbf{Generalization to Larger-Scale Sudoku.} Accuracy (\%) on 9$\times$9 Sudoku based on LLaDA-8B-Instruct. ${\dagger}$ indicates our re-implementation.}
\label{table:sudoku9_llada}
\end{table}

\begin{figure}[t]
\centering
\begin{subfigure}{0.48\columnwidth}
  \centering
  \includegraphics[width=\linewidth]{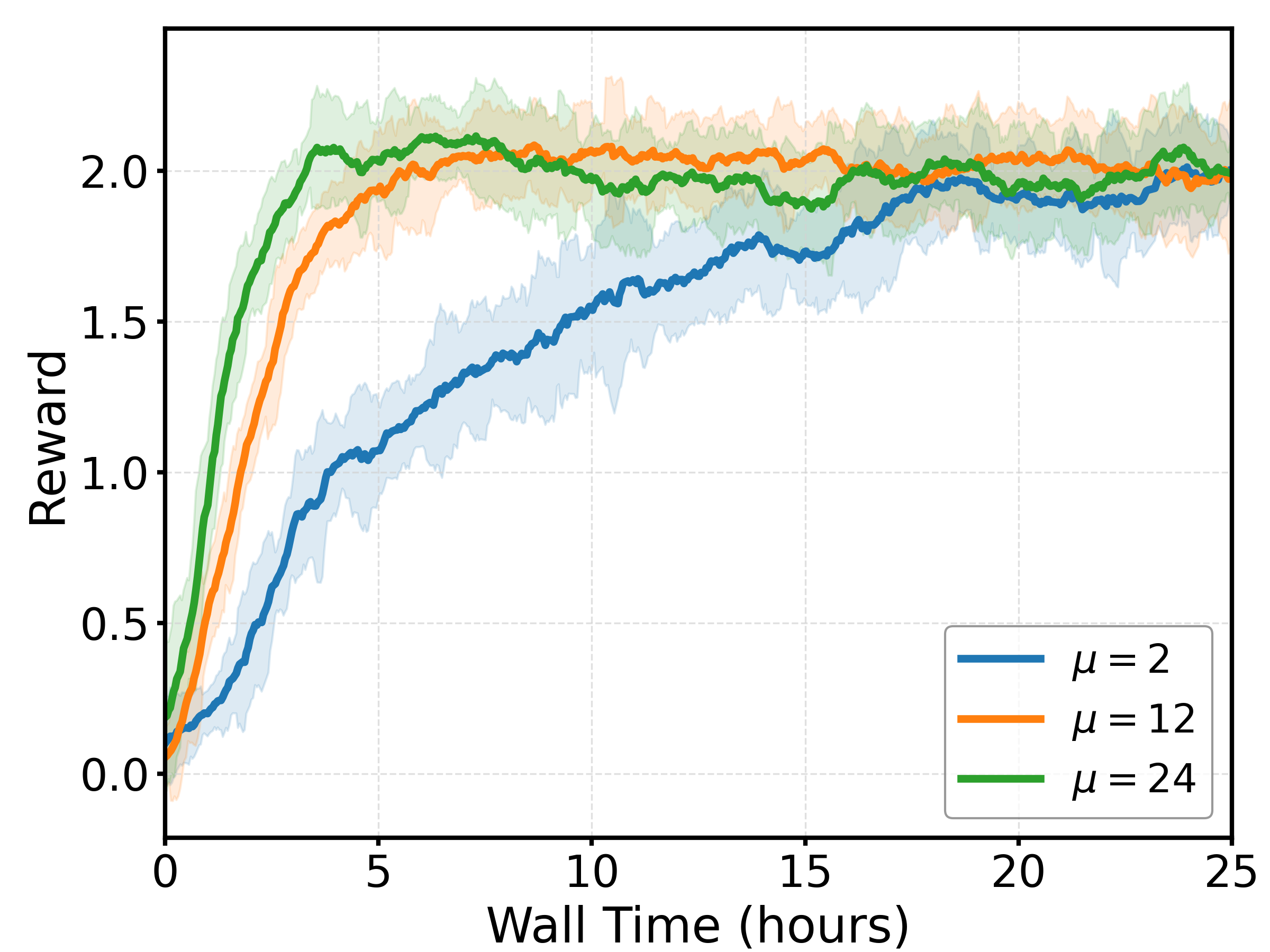}
  \caption{Impact of $\mu$}
  \label{fig:mu_impact}
\end{subfigure}
\hfill
\begin{subfigure}{0.48\columnwidth}
  \centering
  \includegraphics[width=\linewidth]{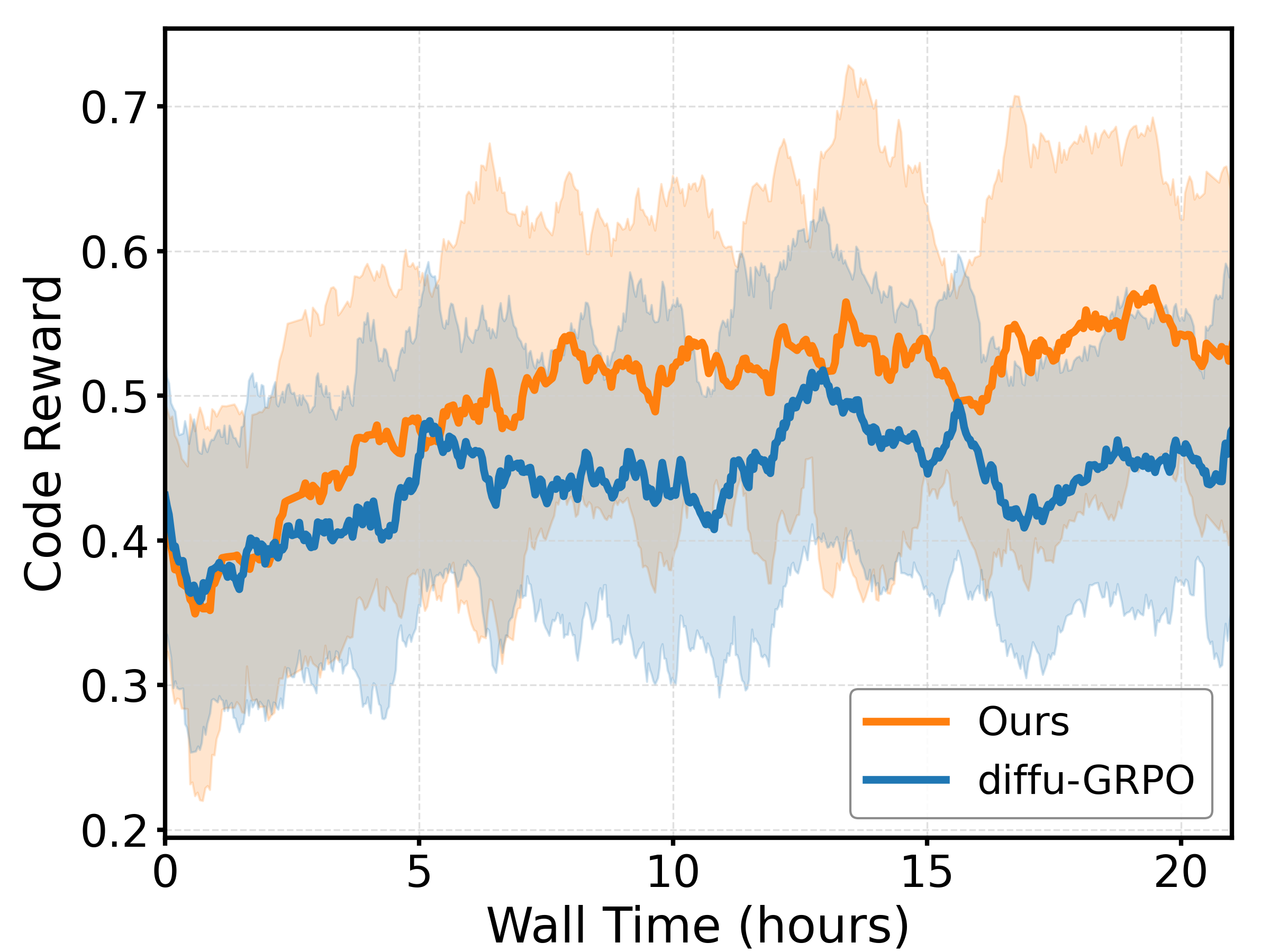}
  \caption{Efficiency on code tasks}
  \label{fig:code_efficiency}
\end{subfigure}

\caption{\textbf{Training Efficiency Analysis.} (a) Impact of policy update steps $\mu$ on reward convergence. (b) Reward convergence comparison on code generation tasks under the same GPU budgets.}
\label{fig:efficiency_analysis}
\end{figure}

\subsection{Efficiency}
A practical concern is whether step-aware process rewards computation introduces prohibitive overhead. We analyze training efficiency along two dimensions.
Firstly, Figure~\ref{fig:efficiency_analysis}\,(a) shows how the number of policy optimization updates per rollout batch $\mu$ affects sample efficiency and stability. Raising $\mu$ from 2 to 24 accelerates reward convergence under the same wall time, but $\mu=24$ becomes unstable late in training and the reward declines. With $\mu=12$, convergence remains stable, reaches a higher final reward with faster wall-clock convergence.
Secondly, relative to diffu-GRPO, PAPO incurs additional per-step overhead from process reward computation. However, even in the most expensive setting of code generation with execution-based rewards (Figure~\ref{fig:efficiency_analysis}\,(b)), PAPO still reaches a higher reward in less wall time despite the extra cost per step. This result indicates that the process-aligned learning signal more than compensates for the per-step overhead and improves overall training efficiency. A broader efficiency analysis across all tasks is provided in Appendix~\ref{B}.

\section{Conclusion}
In this paper, we identify two fundamental misalignments when applying reinforcement learning to diffusion large language models (dLLMs): process-reward misalignment due to the reliance on sparse, terminal rewards, and inefficient state selection caused by training on unfaithful contexts with uniform step sampling. To address these limitations, we propose Process Aligned Policy Optimization (PAPO), a novel framework that holistically aligns the RL update process with the model's authentic generative trajectory. At its core, PAPO features two synergistic modules: Step-Aware Process Rewards (SPR) for dense, step-specific credit assignment, and Entropy-Guided Historical Re-enactment (EHR) for efficient, authentic state selection. Extensive experiments on four reasoning benchmarks demonstrate that PAPO significantly outperforms strong baselines. We believe our process aligned approach offers a more stable and efficient path to enhancing the complex reasoning capabilities of dLLMs.

\clearpage
\section*{Limitations} While PAPO achieves strong performance on text-based reasoning, two extensions remain open. Process reward relies on a one-step denoised prediction $\hat{x}_0(t)$ under multi-step mask diffusion rollout. As few-step distillation (e.g. T3D~\cite{few-step}), and training-free fast decoders (e.g. Fast-dLLM~\cite{fast-dllm}) reduce the number of rollout steps, it will be valuable to define process rewards at the block level or with limited multi-step lookahead, so as to improve reward fidelity without prohibitive overhead. Moreover, the current experiments are conducted exclusively on text-based reasoning benchmarks. Further studies are needed to evaluate the effectiveness of process-aligned RL on multimodal dLLMs~\cite{yang2025mmada,li2026omnidiffusionunifiedmultimodalunderstanding,xin2025luminadimooomnidiffusionlarge}, where modality-specific process rewards and domain structures may differ substantially from the text setting.

\section*{Acknowledgments} This work is supported by the National Natural Science Foundation of China (NSFC) under Grants 62306295, 62576328, and 625B2175. The AI-driven experiments, simulations and model training were performed on the robotic AI-Scientist platform of Chinese Academy of Sciences.


\bibliography{custom}

\appendix

\clearpage
\section{ Implementation Details}
\label{A}
\addcontentsline{toc}{section}{\textcolor[rgb]{0,0,0}{A. Implementation Details}}

\subsection{Reward Functions}
\label{A.1}
\addcontentsline{toc}{subsection}{\textcolor[rgb]{0,0,0}{A.1. Reward Functions}}
Following d1~\cite{zhao2025d1}, we utilize a task-specific, composite reward function to provide granular feedback during the reinforcement learning phase, which ensures the model is rewarded not only for reaching the correct final answer but also for generating structurally sound and coherent reasoning steps. The rewards are detailed below.

\textbf{GSM8K}. For the GSM8K dataset, we apply a composite reward function composed of five distinct components. 
(1) XML Structure Reward: This component validates the fundamental tag-based structure. A reward of +0.125 is granted for each correctly placed opening and closing tags. Additionally, a minor penalty is applied for any extraneous content appearing after the final tag to ensure the output terminates cleanly.
(2) Soft Format Reward: A reward of +0.5 is awarded if the entire response matches the general pattern: <reasoning>...</reasoning><answer>...</answer>.
(3) Strict Format Reward: A reward of +0.5 is given for adherence to the exact prescribed format with correct line breaks.
(4) Integer Answer Reward: A reward of +0.5 is provided if the answer is a valid integer. 
(5) Correctness Reward: A reward of +2.0 is granted if the extracted answer exactly matches the ground-truth. 

\textbf{MATH500}. Similar to GSM8K, the reward for this more advanced math dataset is a composite of structural and accuracy-based rewards:
(1) Formatting Reward: A tiered reward is assigned based on the presence and correct usage of <answer> tags and the $\setminus \mathrm{boxed}$.   A reward of 1.00 for <answer> with $\setminus \mathrm{boxed}$
inside; +0.75 for <answer> without $\setminus \mathrm{boxed}$; +0.50 for $\setminus \mathrm{boxed}$ only. +0.25 for neither. 
(2) Correctness Reward: A  reward of 2.0 is granted if the correct answer is in $\setminus \mathrm{boxed}  \left \{  \right \} $.

\textbf{Countdown}. For this arithmetic planning task, the reward function evaluates the validity and correctness of the generated mathematical expression:
(1) A full reward of 1.0 is given if the expression correctly uses all specified numbers to reach the target value.
(2) A partial reward of 0.1 is given if the expression uses the correct numbers but fails to compute the target value, encouraging the use of correct operands.
(3) A reward of 0 is given in all other cases.

\textbf{Sudoku}. In the 4x4 Sudoku puzzles, the reward directly measures the model's constraint-solving capability. The reward is calculated as the proportion of correctly filled digits within the cells that were initially empty in the puzzle prompt. This metric focuses evaluation on the model's problem-solving ability rather than its capacity to replicate the provided clues.

\subsection{ Training Configurations}
\label{A.2}
\addcontentsline{toc}{subsection}{\textcolor[rgb]{0,0,0}{A.2. Training Configurations}}
All experiments are conducted on 8 NVIDIA A100-80G GPUs, with the same  hyperparameters as d1~\cite{zhao2025d1}: sequences of 256 tokens, batch size of 6 per GPU, policy optimization updates value $\mu =12$ and gradient accumulation over 2 steps. We employ Low-Rank Adaptation (LoRA) with a rank of 128 and a scaling factor of 64. We optimize the model using the AdamW optimizer, with parameters $\beta_1 = 0.9$, $\beta_2 = 0.99$, weight decay of 0.1, learning rate of $3\times 10^{-6}$, and gradient clipping at 0.2. We employ entropy-weighting hyperparameter $\alpha =1$. We train 8000 steps (number of gradient updates) for GSM8K, MATH500 and Countdown. For Sudoku, we train on synthetic generated datasets for 6000 steps. 

\subsection{ Sampling and Evaluation}
\label{A.3}
\addcontentsline{toc}{subsection}{\textcolor[rgb]{0,0,0}{A.3. Sampling and Evaluation}}
Our sampling process follows the semi-autoregressive approach introduced in LLaDA~\cite{llada}. The sequence is first partitioned into multiple blocks that are generated in a left-to-right fashion. During the generation of each individual block, the low-confidence remasking strategy is then employed. Following the practice in d1~\cite{zhao2025d1} and TSE~\cite{wang2025time}, we evaluate the model every 100 steps, starting from step 600 and report the best results.

\begin{figure*}[t]
\centering  
    \begin{overpic}[width=1\linewidth]{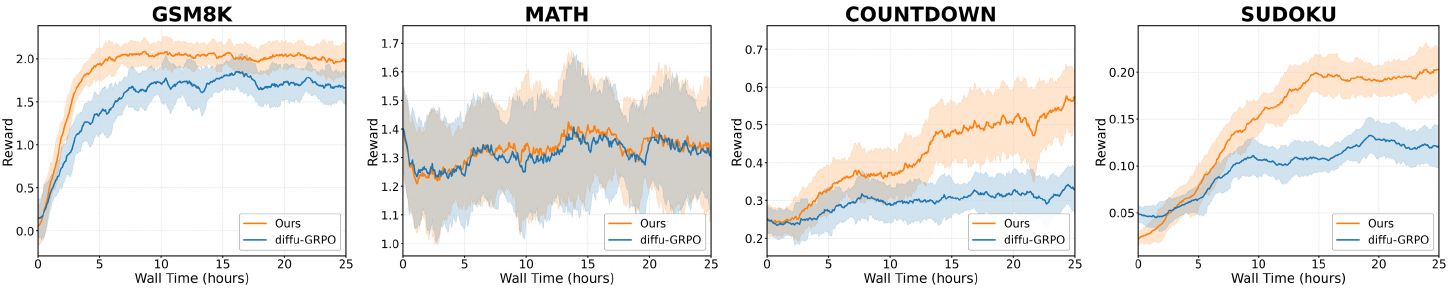} 
    \end{overpic} 
    \vspace{-6mm}
    \caption{\textbf{Training Efficiency and Reward Convergence over Wall Time.} Comparison of reward trajectories for PAPO and the diffu-GRPO baseline across four benchmarks. The results illustrate PAPO's superior training efficiency, as it consistently converges to a higher reward in significantly less wall time.} 
    \label{fig:supp_time}
\end{figure*}

\begin{figure*}[t]
\centering  
    \begin{overpic}[width=1\linewidth]{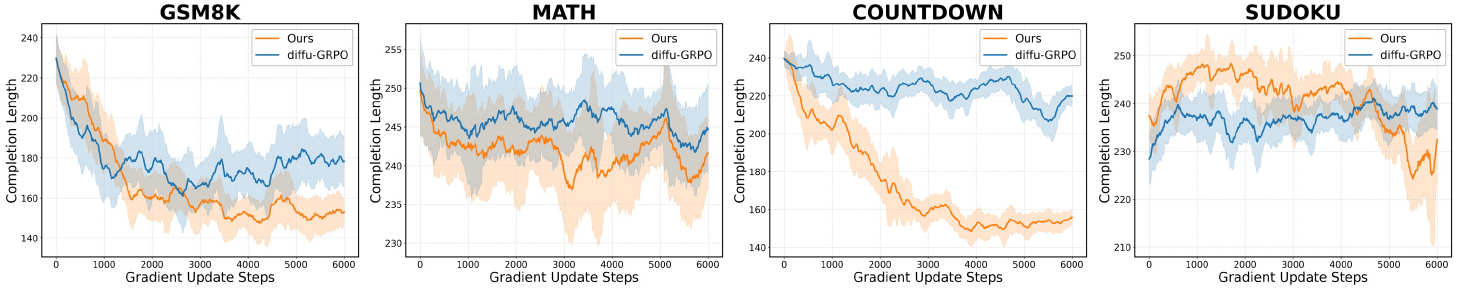} 
    \end{overpic} 
    \vspace{-6mm}
    \caption{\textbf{Dynamics of Completion Length and Token Efficiency.} Evolution of average completion length during training across four benchmarks. PAPO converges to more concise outputs, demonstrating superior token efficiency while achieving higher rewards.} 
    \label{fig:supp_length}
\end{figure*}

\begin{figure*}[!t]
\centering  
    \begin{overpic}[width=1\linewidth]{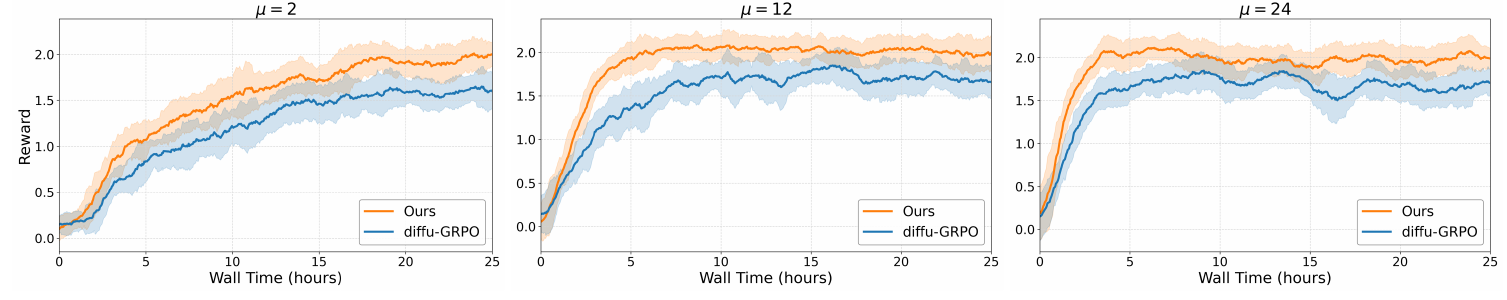} 
    \end{overpic} 
    \vspace{-6mm}
    \caption{\textbf{Impact of Policy Optimization Updates  Values $\mu$ on Performance and Stability.} PAPO demonstrates superior performance compared to diffu-GRPO across various values of $\mu$. This advantage is particularly evident at $\mu=24$, where our approach maintains robust stability and showcases excellent scalability, in contrast to the baseline which becomes notably unstable.} 
    \label{fig:u}
\end{figure*}

\section{Additional Experiments Analysis}
\label{B}
\addcontentsline{toc}{section}{\textcolor[rgb]{0,0,0}{B. Additional Experiments Analysis}}

\subsection{ Training and Computational Costs}
\label{B.1}
\addcontentsline{toc}{subsection}{\textcolor[rgb]{0,0,0}{B.1. Training and Computational Costs}}
To evaluate its practical efficiency, we compare the training time of PAPO against the diffu-GRPO baseline. On the surface, PAPO introduces a slight computational overhead by calculating a process reward for each of the policy update steps $\mu$, guided by our entropy-based state selection. Crucially, this overhead is strictly bounded, as process rewards are calculated only for the specific $\mu$ historical states re-enacted via our entropy-guided selection, not for all timesteps in the trajectory. However, this minor addition is compensated by the elimination of a costly step inherent in diffu-GRPO: the creation of artificial training inputs through prompt masking or completion masking for each update. Instead, PAPO efficiently re-enacts authentic, pre-cached historical states, resulting in a lower wall time per training step. As shown in Figure~\ref{fig:supp_time}, this per-step efficiency translates directly into superior training dynamics. PAPO not only converges to a higher reward but also achieves this in significantly less time, a marked advantage in both speed and final performance that is consistently demonstrated across all reasoning and planning benchmarks.

\subsection{ Completion Lengths Dynamics Analysis}
\label{B.2}
\addcontentsline{toc}{subsection}{\textcolor[rgb]{0,0,0}{B.2. Completion Lengths Dynamics Analysis}}
We further analyze the token efficiency of our framework by tracking the average completion length during training, as shown in Figure~\ref{fig:supp_length}. PAPO demonstrates a clear advantage on the GSM8K, Math and Countdown benchmarks, converging to significantly shorter output sequences while simultaneously achieving superior rewards. A similar trend towards more concise solutions is observed for Sudoku, where the policy learns to progressively shorten its completion length during training. This consistent reduction in completion length indicates that PAPO learns not only a more effective but also a more token-efficient reasoning policy than the baseline.

\subsection{Impact of Policy Optimization Update Values $\mu$} 
We investigate the impact of policy optimization update values $\mu$ on both sample efficiency and stability, with results shown in Figure~\ref{fig:u}. Beyond consistently outperforming diffu-GRPO, PAPO demonstrates excellent scalability by effectively leveraging a larger $\mu$ to perform more policy updates from each data batch, which translates into a steeper learning curve and faster wall time convergence. The superior scalability and robustness of PAPO become most apparent under aggressive update schedules. At $\mu=24$, PAPO maintains a stable, high-reward trajectory while diffu-GRPO begins to exhibit training instability, evidenced by reward volatility. This confirms that the rich, stable learning signals from PAPO's synergistic modules enable more robust and efficient policy optimization.

\section{ More Discussion of PAPO and Future Work}
\label{C}
\addcontentsline{toc}{section}{\textcolor[rgb]{0,0,0}{C. More Discussion of PAPO and Future Work}}
While PAPO demonstrates significant improvements in reasoning for diffusion language models, we identify several promising avenues for future work based on current design. 

\subsection{  Generalization to Diverse Architectures and Modalities}
\label{C.1}
\addcontentsline{toc}{subsection}{\textcolor[rgb]{0,0,0}{C.1. Generalization to Diverse Architectures and Modalities}}
Our current work validates the effectiveness of PAPO on text-based reasoning tasks based on a full-attention dLLM architecture~\cite{llada}. However, the core principle of process alignment is not inherently limited to this specific setup. With the rise of alternative architectures like block-diffusion models (e.g. SDAR~\cite{sdar}) and the expansion of dLLMs into multimodal domains (e.g. MMaDA~\cite{yang2025mmada}, Lumina-DiMOO~\cite{xin2025luminadimooomnidiffusionlarge}), a key future direction is to adapt and evaluate PAPO on these emerging platforms. Extending our process aligned RL framework to enhance reasoning in these new contexts represents a significant next step.

\begin{figure*}[t]
\centering  
    \begin{overpic}[width=1\linewidth]{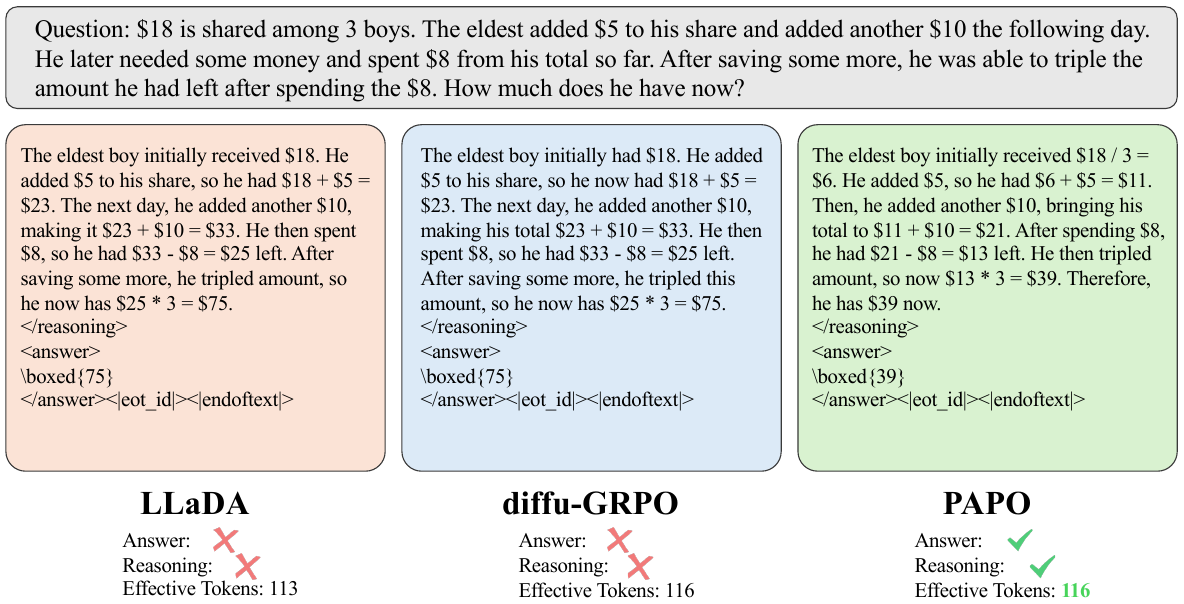} 
    \end{overpic} 
    \vspace{-6mm}
    \caption{\textbf{Comparison of Generated Responses on the GSM8K Benchmark.}} 
    \label{fig:supp_gsm8k}
\end{figure*}
\begin{figure*}[!t]
\centering  
    \begin{overpic}[width=1\linewidth]{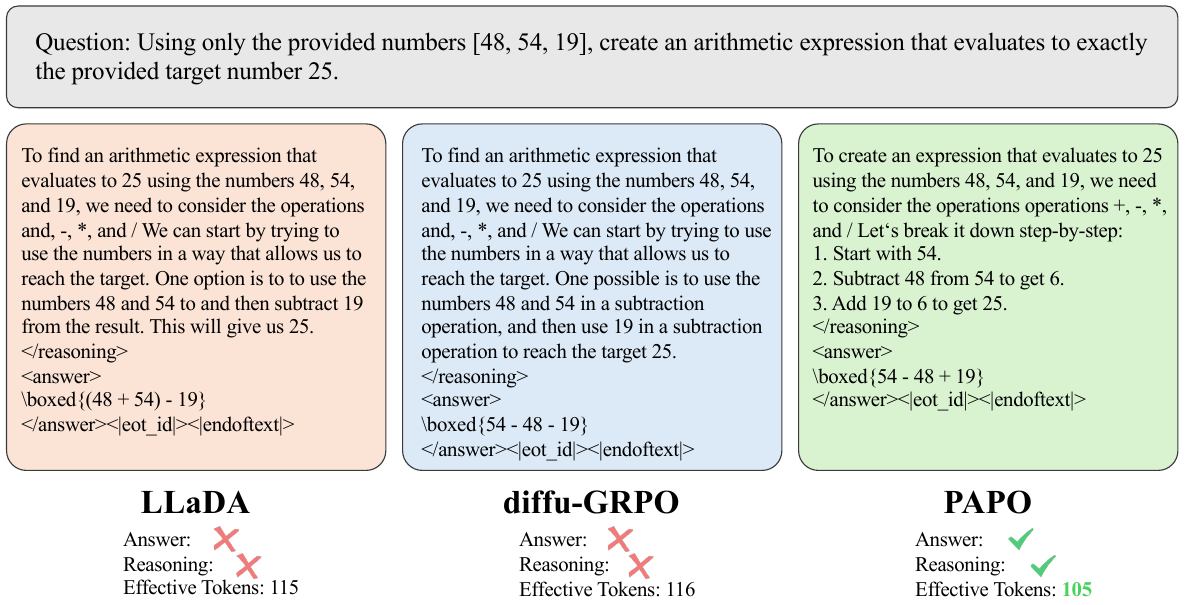} 
    \end{overpic} 
    \vspace{-6mm}
    \caption{\textbf{Comparison of Generated Responses on the Countdown Benchmark.}} 
    \label{fig:supp_countdown}
\end{figure*}
\subsection{ Computation and Memory Efficiency Analysis}
\label{C.2}
\addcontentsline{toc}{subsection}{\textcolor[rgb]{0,0,0}{C.2. Computation and Memory Efficiency Analysis}}
At first glance, PAPO might appear to introduce significant overhead due to the computation of process rewards and the storage of historical rollout states. However, our implementation is highly efficient and designed to minimize this cost. We do not compute process rewards or store states for all $T$ denoising steps. Instead, we first leverage entropy-guided sampling to select the $\mu$ most informative steps for policy updates. Only then do we cache the corresponding historical states $x_t$ and compute their process rewards through $\hat{x}_0{(t)}$. Crucially, as illustrated in Figure~\ref{fig:method_rollout} in the main paper, the state $\hat{x}_0{(t)}$ required for calculating the process reward is a transient product of the standard rollout process, generated between the denoise and remask operations. This means we do not need any extra generation steps to obtain these states, making the additional cost minimal. As shown in Figure~\ref{fig:supp_time}, our method not only avoids significant overhead but leads to faster overall convergence, ultimately achieving superior performance.

\subsection{ Fidelity vs. Efficiency in Process Rewards}
\label{C.3}
\addcontentsline{toc}{subsection}{\textcolor[rgb]{0,0,0}{C.3. Fidelity vs. Efficiency in Process Rewards}} 
The efficacy of our SPR mechanism relies on the one-step denoised prediction $\hat{x}_0{(t)}$ as an efficient proxy for intermediate reasoning quality. This design prioritizes computational efficiency over complex and costly alternatives like training auxiliary reward  model or value model. The effectiveness of this lightweight approach is empirically validated by our ablation studies (Table~\ref{table:ablation} in the main paper) and the clear upward trend of process rewards during training (Figure~\ref{fig:pr} in the main paper), which confirms that SPR provides timely and effective step-wise feedback. Nonetheless, the success of this approach is contingent on the ability of the one-step reward to faithfully reflect the quality of the current action. While our experiments (Table~\ref{table:multi-step} in the main paper) confirm that one-step prediction offers the best fidelity-efficiency trade-off under standard multi-step rollouts, emerging few-step generation paradigms such as T3D~\cite{few-step} and Fast-dLLM~\cite{fast-dllm} substantially reduce the number of rollout steps and the associated computational cost, fundamentally changing the trajectory structure. Adapting SPR to these compressed trajectories, for instance by defining process rewards at the block level or calibrating reward granularity to the reduced trajectory length, remains a promising direction for future work.


\section{ Qualitative Examples}
\label{D}
\addcontentsline{toc}{section}{\textcolor[rgb]{0,0,0}{D. Qualitative Examples}}
Figure~\ref{fig:supp_gsm8k} and Figure~\ref{fig:supp_countdown} provide qualitative examples from the math reasoning (GSM8K) and planning (Countdown) benchmarks respectively, which highlight the superior reasoning capabilities of our PAPO framework. In the GSM8K example (Figure~\ref{fig:supp_gsm8k}), PAPO correctly interprets the problem's sequential logic by accurately accounting for all conditions, whereas baseline models like diffu-GRPO falter by overlooking critical initial constraints, resulting in flawed reasoning paths. Similarly, on the Countdown benchmark (Figure~\ref{fig:supp_countdown}), PAPO generates a more structured, step-by-step reasoning path that is not only logically sound but also more token-efficient than the outputs from LLaDA and diffu-GRPO. This ability to maintain a coherent and concise reasoning trace is directly attributable to our Step-Aware Process Rewards (SPR), which ensure that each intermediate step's contribution is properly valued and reinforced. Furthermore, our Entropy-Guided Historical Re-enactment (EHR) equips the policy with authentic and informative reasoning states, facilitating its learning of efficient and accurate problem-solving strategies.
\end{document}